\documentclass[12pt]{article}
\usepackage{graphicx,color}
\usepackage{amsmath}
\usepackage{geometry}

\def\sko {\vspace{.1in}}

\def\C {\,|\:}

\newcommand{\bi}{\begin{itemize}}
\newcommand{\ib}{\end{itemize}}

\newcommand{\be}{\begin{enumerate}[(i)]}
\newcommand{\eb}{\end{enumerate}}

\title{
\Large Fully Nonparametric Bayesian Additive Regression Trees
}
\author{%
Edward George \and Prakash Laud \and Brent Logan \and Robert McCulloch \and Rodney Sparapani 
\footnote{%
Edward I. George, Department of Statistics, The Wharton School, University of Pennsylvania.  
Prakash Laud, Division of Biostatistics, Medical College of Wisconsin.
Brent Logan, Division of Biostatistics, Medical College of Wisconsin.
Robert E. McCulloch, School of Mathematical and Statistical Sciences, Arizona State University,
Robert.McCulloch@asu.edu.
Rodney Sparapani, Division of Biostatistics, Medical College of Wisconsin.
}
}

\begin{document}

\maketitle
\thispagestyle{empty}

\begin{abstract}
\noindent 
Bayesian Additive Regression Trees (BART) is a fully Bayesian approach to
modeling with ensembles of trees.
BART can uncover complex regression functions with high dimensional
regressors in a fairly automatic way and provide Bayesian quantification
of the uncertainty through the posterior.
However, BART assumes IID normal errors.
This strong parametric assumption can lead to misleading
inference and uncertainty quantification.
In this paper, we use the classic Dirichlet process mixture (DPM) mechanism to 
nonparametrically model the error distribution.
A key strength of BART is that default prior settings work
reasonably well in a variety of problems.
The challenge in extending BART is to choose the parameters of the DPM
so that the strengths of the standard BART approach is not lost
when the errors are close to normal, but the DPM has the ability
to adapt to non-normal errors.

\end{abstract}

\newpage

\setcounter{page}{0}
\tableofcontents
\thispagestyle{empty}

\newpage

\section{Introduction}\label{sec:introduction}

Data analysts have long sought to uncover the information 
in data without making strong assumptions about the nature of the underlying process.
Traditionally, asymptotic approaches to frequentist inference have been used.
The asymptotics often play the role of minimizing the assumptions needed to make an inference.
As an alternative to frequentist reasoning, Bayesian methods have been suggested as a more
coherent approach to uncertainty quantification.
Bayesian approaches do not require assumptions about the small sample validity of the asymptotic
approximation, but they do require the specification of a full probability model.
For a particular cogent presentation of both viewpoints see Poirier 1995 \cite{Poirier95}.

On another front, a whole set of new tools for dealing with high-dimensional and ``big'' data
have recently been developed.
Varian 2014 \cite{Var14} surveys what he thinks are the more important developments.
Varian says ``I believe that these methods have a lot to offer and should be more widely
known and used by economists. In fact, my standard advice to graduate students
these days is go to the computer science department and take a class in machine
learning.''
Prominently featured in Varian's article are methods based on binary decision trees including boosting 
and Random Forests.

In this article we suggest that BART, and more precisely the variant developed in this paper: DPMBART
should be considered.
DPMBART builds upon both the modern Machine Learning approaches and fundamental Bayesian technology.
BART stands for Bayesian Additive Regression Trees, and was introduced in
Chipman, George, and McCulloch 2010 {\cite{chipman:etal:2010}.
BART is most closely related to boosting in that it combines a large set of relatively
simple decision trees to fit a complex high-dimensional response.
BART fits the model $Y = f(x) + \epsilon$ while making minimal assumptions about $f$.
The function $f$ is represented as the sum of many trees. 
In this paper we extend BART by using Bayesian nonparametrics (Dirichlet process mixtures, henceforth DPM) to 
model the error terms.  This model we dub DPMBART.  
Our hope is that DPMBART will work in a wide variety of applications with minimal assumptions
and minimal tuning.  
In order to avoid the need to tune, much of our effort is devoted to the specification of
a data based prior.
While data-based priors are not strictly ``Bayesian'' recent work 
(Yushu Shi and Michael Martens and Anjishnu Banerjee and Purushottam Laud} (2017) \cite{ShiMart17})
have shown them to be particularly effective in the often tricky specification of priors for DPM models.
In addition, much of the success of BART is due to a relatively simple data based prior.
While we believe that inclusion of prior information should play a role in most analyses,
we also want a tool that can give good results simply.

Our hope is that DPMBART harnesses the power of the recent developments discussed by Varian
and combines them with modern Bayesian methodology
to provide a viable solution to the basic problem of obtaining a reasonable inference in high dimensions
with minimal assumptions.

This paper is organized as follows.
In Section~\ref{sec:bart}, we review the standard BART model.
In Section~\ref{sec:dpmbart}, we combine the BART and DPM technologies to 
construction a fully nonparametric version of BART: DPMBART.
The key is the choice of priors for the DPM part of the model.
In Section~\ref{sec:examples}, we present examples to illustrate the performance
of our DPMBART model.
While an advantage of BART is that prior information can be used, our emphasis
is on the performance of default prior settings in a variety of real and simulated examples.
Section~\ref{sec:conclusion} concludes the paper.

\newpage
\section{BART: Bayesian Additive Regression Trees}\label{sec:bart}

In this section we review the basic BART model.
is a fully Bayesian development of an ensemble of trees model in which the overall
conditional mean of a response given predictors is expressed as the sum of many trees.
The BART algorithm includes an effective Markov Chain Monte Carlo algorithm
which explores the complex space of an ensemble of trees without prespecifying the dimension
of each tree.  Guided by the prior, the complexity of the model is inferred.
BART is motivate by the fundamental work of Friedman \cite{Fri:2001}.
The seminal work is due to Freund and Schapire \cite{Fre:Sch:1997}.
Following Freund and Schapire, these types of ensemble models and associated algorithms are often referred to as
``boosting'' algorithms.  The overall model is made up of many small contributions from ensemble models.
Typically, as in BART, the models in the ensemble are binary trees.

Let $y$ denote the response and $x$ denote the vector of predictor variables.
BART consider the basic model
\begin{equation}\label{eqn:basicmodel}
Y_i = f(x_i) + \epsilon_i, \;\; \epsilon_i \sim N(0,\sigma^2).
\end{equation}
The goal is to be able to infer a the function $f$ with minimal assumptions and
high dimensional $x$.
In the spirit of boosting, BART lets
\begin{equation}\label{eqn:sumoftrees}
f(x) = \sum_{j=1}^m g(x; T_j,M_j)
\end{equation}
where each $g(x; T_j,M_j)$ represents the function captured by a single binary tree.
Each binary tree is described by the tree $T$ which encode the structure of the tree and all
of the decision rules and $M = (\mu_1,\mu_2,\ldots,\mu_b)$ which records the values associated
with each bottom or leaf node of the tree which has $b$ bottom nodes.  

\begin{figure}
\centerline{\includegraphics[scale=.25]{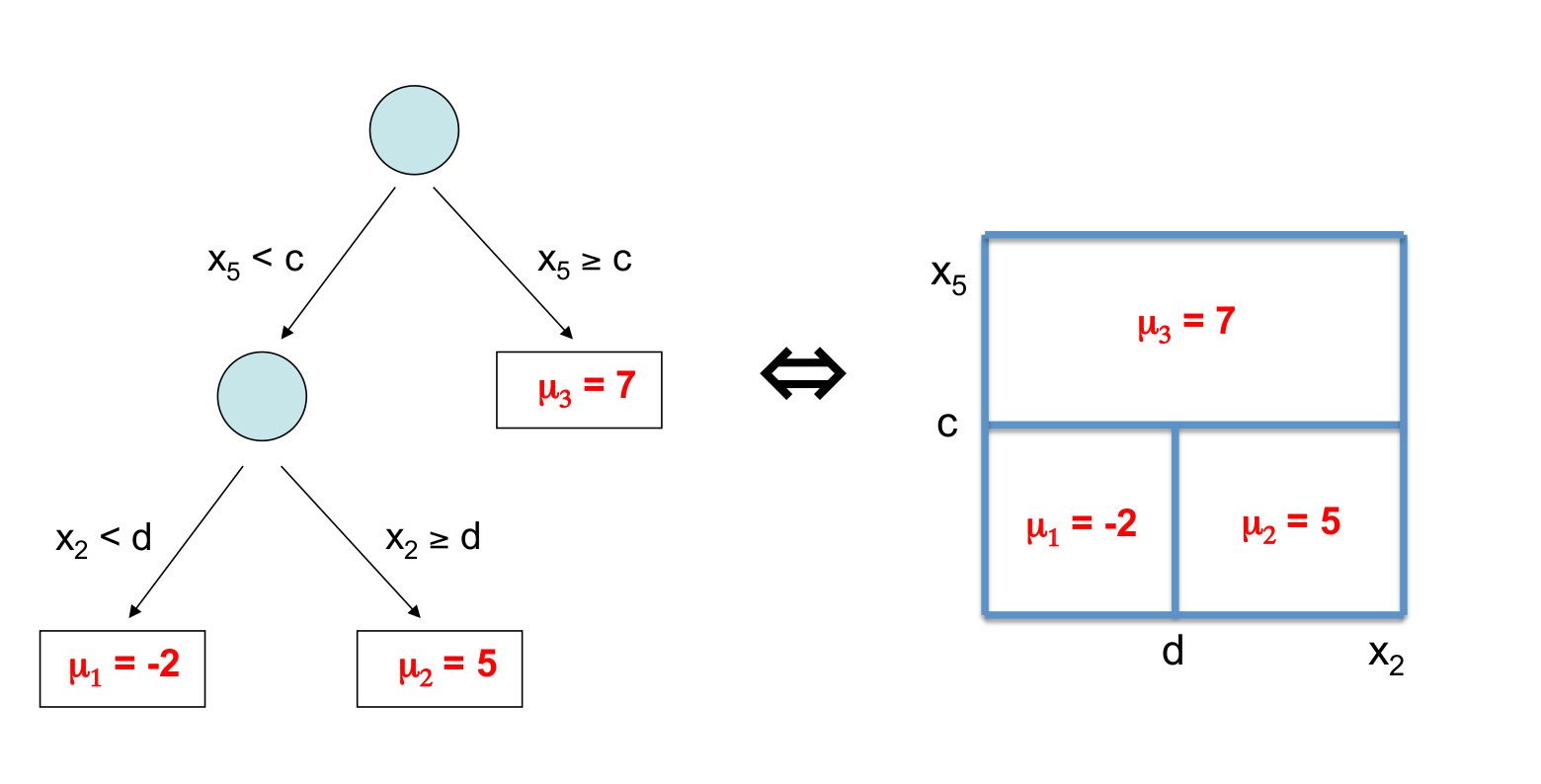}}
\caption{%
A single binary tree.
In left panel is the binary tree.
In the right panel is the corresponding partition of the predictor space giving
a step function.  Each region is labeled with the corresponding mean level taken
from the leaf nodes of the tree in the left panel.
\label{fig:single-tree}}
\end{figure}

Figure~\ref{fig:single-tree} depicts a single tree $g(x; T, M)$.
The tree has two interior nodes with decision rules of the form $x_j < c$ meaning if the $j^{th}$
coordinate of $x$ is less than $c$ then you are sent left, otherwise you go right.
The function $g(x; T, M)$ is evaluated by dropping $x$ down the tree.
At each interior decision node, $x$ is sent left or right until it hits a bottom node
which contains a $\mu$ value which is then the returned value of $g$.
In Figure~\ref{fig:single-tree} we have $b=3$ bottom nodes.
In Figure~\ref{fig:single-tree} the tree only uses $x_2$ and $x_5$, but $x$ could contain many
more predictors.

Single trees play a central role in modern statistics.
However, as seen in Figure~\ref{fig:single-tree}, they represent a rather crude function.
The boosting ensemble approach represented in Equation~\ref{eqn:sumoftrees} has the 
ability to accurately capture high dimensional functions.

\subsection{BART Prior}

BART entails both a prior on the parameter
$\Theta = ((T_1,M_1),\ldots,(T_m,M_m), \sigma)$
and a MCMC algorithm for exploring the posterior.
Note that the dimension of each $T_j$ is not fixed.

The prior has the form
$$
p(\Theta) = p(\sigma) \, \prod_{j=1}^m \, p(T_j,M_j)
$$
with $p(T,M) = p(T) \, p(M \C T)$.
Note that the dimension of $M$ depends on $T$.

$$
p(M \C T) = \prod_{i=1}^m \, p(\mu_i)
$$
with $p(\mu) \sim N(0,\tau^2)$ so that conditional on all the trees all the $\mu$'s at the
bottom of all three are iid $N(0,\tau^2)$.
The prior for $\sigma$ is the standard inverted chi-squared: $\sigma^2 \sim (\nu \lambda)/(\chi^2_\nu)$.
Chipman et al. describe a tree growing process to specify  the prior $p(T)$ and data based default prior
choices for $\tau$ and $(\nu,\lambda)$.
We refer the reader to Chipman et al. for details but discuss the choice of prior 
for $(\nu,\lambda)$ in detail in Section~\ref{subsec:priorsigma} as this is needed for the development of this paper.
All other details may be left in the background.
The key to these prior choices is that the prior expresses of preference for small trees and shrinks
all the $\mu$ towards zero in such a way that only the overall sum can capture $f$.
These prior specifications enable BART to make each individual tree a ``weak learner'' in
that it only makes a small contribution to the overall fit.

\subsection{BART MCMC}

Given the observed data $y$, the BART model induces a
posterior distribution
\begin{equation}
p((T_1,M_1), \ldots,(T_m,M_m),\sigma | \,y) \label{posterior}
\end{equation}
on all the unknowns that determine a sum-of-trees model
(\ref{eqn:basicmodel} and \ref{eqn:sumoftrees}).  Although the sheer size of the parameter
space precludes exhaustive calculation,  the following backfitting
MCMC algorithm can be used to sample from this posterior.

At a general level, the algorithm is a Gibbs sampler.  For
notational convenience, let $T_{(j)}$ be the set of all trees in
the sum {\em except} $T_j$, and similarly define $M_{(j)}$. Thus
$T_{(j)}$ will be a set of $m-1$ trees, and $M_{(j)}$ the
associated terminal node parameters.  The Gibbs sampler here
entails $m$ successive draws of $(T_j,M_j)$ conditionally on
$(T_{(j)}, M_{(j)}, \sigma)$:
\begin{equation}
(T_j,M_j) | T_{(j)}, M_{(j)}, \sigma, y,  \label{draw1}
\end{equation}
$j = 1,\ldots,m$, followed by a draw of $\sigma$ from the full conditional:
\begin{equation}
\sigma | T_1,  \ldots T_m, M_1, \ldots, M_m, y \label{draw2} .
\end{equation}

Each $(T,M)$ draw in \ref{draw1} is done by letting
$p(T,M \C \circ) = p(T \C \circ) \, p(M \C T,\circ)$
where $\circ$ denotes all the other conditioning information.
Given the prior choices we can analytically integrate out the $\mu$ to obtain
an computationally convenient expression for $p(T \C \circ)$
Metropolis Hastings steps are then use to propose changes to $T$.
While many useful steps are in the literature the key steps are the
birth/death pair.  A birth step proposes adding a decision rule to a bottom node
of the current tree so that it spawns left and right child bottom nodes.
A death move proposes the elimination of a left/right pair of bottom nodes.
This key birth/death pair of moves allows the MCMC to explore trees of varying complexity and size.

\subsection{Specification of the prior on  $\sigma$}\label{subsec:priorsigma}

For $p(\sigma)$, we use the (conditionally) conjugate inverse chi-square distribution
$\sigma^2 \sim \nu \, \lambda/\chi_{\nu}^2$.  To guide the specification
of the hyperparameters $\nu$ and $\lambda$, we recommend a data-informed approach in order to assign substantial probability to the entire region of
plausible $\sigma$ values while avoiding overconcentration and overdispersion.
This entails calibrating the prior degrees of freedom $\nu$ and scale $\lambda$
using a ``rough data-based overestimate''
$\hat{\sigma}$ of $\sigma$.

The two natural choices for $\hat{\sigma}$ are (1) the ``naive''
specification, in which we take $\hat{\sigma}$ to be the sample
standard deviation of $Y$ (or some fraction of it), or (2) the ``linear model''
specification, in which we take $\hat{\sigma}$ as the residual
standard deviation from a least squares linear regression of $Y$
on the original $x$'s.  We then pick a value of
$\nu$ between 3 and 10 to get an appropriate shape, and a value of
$\lambda$ so that the $q$th quantile of the prior on $\sigma$ is located at
$\hat{\sigma}$, that is $P(\sigma < \hat{\sigma}) = q.$  We consider values of $q$ such as
0.75, 0.90 or 0.99 to center the distribution below $\hat{\sigma}$.
For automatic use, we recommend the
default setting $(\nu, q) = (3, 0.90)$ which tends to avoid
extremes.  Alternatively, the values of $(\nu, q)$ may be chosen by
cross-validation from a range of
reasonable choices.

The dashed density in Figure~\ref{fig:baseprior-lambda} shows the prior for $\sigma$
with $\hat{\sigma}=1.0$

\newpage

\newpage
\section{DPMBART: Fully Nonparametric Bayesian Additive Regression Trees}\label{sec:dpmbart}
In this section we develop our fully nonparametric Dirichlet process mixture
additive regression tree (DPMBART) model.
Our base mode is
$$
Y_i = f(x_i) + \epsilon_i.
$$
As in BART, we use a sum of trees to model the function $f$.
However, unlike BART, we do not want to make the restrictive assumption that the errors are iid normal.
Following Escobar and West, we use a Dirichlet process mixture (DPM) to nonparametrically model the errors.
Excellent recent textbook discussions of this model are in Muller et al. \cite{MQJH15} and Rossi \cite{ROSSI14}.
The key is the specification of the DPM parameters so that it works
well with the BART additive tree structure  giving good performance in a variety of situations
without a lot of tuning.

A simple way to think about the DPM model for the errors is to let
$$
\epsilon_i \sim N(\mu_i,\sigma_i^2),
$$
so that each error $\epsilon_i$  has its own mean $\mu_i$ and standard deviation $\sigma_i$.
Of course, this model 
is too flexible without further structure.
Let $\theta_i = (\mu_i,\sigma_i)$. 
The DPM magic adds a hierarchical model for the set of  $\theta_i$ so that 
there is a random number of unique values. 
Each observation can have its own $\theta$, but observations share $\theta$ values
so that the number of unique values is far less than the sample size.
This reduces the effective complexity of the parameter space.

The DPM hierarchical model draws a discrete distribution using the  Dirichlet process (DP)
and then draws the $\theta_i$ from the discrete distribution.  Because the distribution
is discrete, with positive probability, some of the $\theta_i$ values will be repeats.
Letting $G$ denote the random discrete distribution our hierarchical model is:

$$
\{\epsilon_i\} \C \{\theta_i\}, \;\;  \{\theta_i\}\C G, \;\;  G \C G_0,\alpha
$$

with,

$$
\epsilon_i \sim N(\mu_i,\sigma_i^2), \;\; \theta_i = (\mu_i,\sigma_i) \sim G, \;\; G \sim DP(G_0,\alpha).
$$

where $\{x_i\}$ means $\{x_i\}_{i=1}^n$.
$DP$ denotes the Dirichlet process distribution over discrete distributions given parameters
$G_0$ and $\alpha$.  While we refer the readers to the texts cited above for the details to follow
our prior choices we need some basic intuition about $G_0$ and $\alpha$.
$G_0$ is a distribution over the space of $\theta$.  The atoms of $G$ are iid draws from $G$.
The parameter $\alpha$ determines the distribution of the weights given
to each atom of the discrete $G$.  
A larger $\alpha$ tends to give you a discrete $G$ with more atoms receiving non-negligible weight.
A smaller $\alpha$ means $G$ tends to have most of its mass concentrated on just a few atoms.
In terms of the ``mixture of normals'' interpretation, $G_0$ tells us what normals are likely
(what $\theta = (\mu,\sigma)$ are likely) and $\alpha$ tells us how many normals there are with what weight.

In Section~\ref{subsec:baselineprior} we discuss the form of $G_0$ and approaches for choosing the 
associated parameters.
In Section~\ref{subsec:alphaprior} we discuss the choice of prior for $\alpha$.

\subsection{Specification of Baseline Distribution Parameters}\label{subsec:baselineprior}

Section~\ref{subsec:priorsigma} details the specification of the prior on $\sigma$ in the BART
model with $\epsilon_i \sim N(0,\sigma^2)$.  In this model,  the error distribution is completely
determined by the parameter $\sigma$ so this prior is all that is needed to describe the error
distribution.

Our goal is to specify the family $G_0$ and associated parameters in a way the makes sense
in the DPM setting and captures some of the successful features of the simple BART prior on $\sigma$.
In addition, we need to be aware that if $G_0$ is diffuse, we will tend to get a posterior with few 
unique $\theta$ values,  a feature of the model not evident from the brief description of the DPM model given above.

The $G_0$  family has the commonly employed form, 
$p(\mu,\sigma \C \nu,\lambda,\mu_0,k_0) = p(\sigma \C \nu,\lambda) \, p(\mu \C \sigma,\mu_0,k_0)$ with:
$$
\sigma^2 \sim \frac{\nu \lambda}{\chi^2_\nu}; \;\; \mu \C \sigma \sim N(\mu_0,\frac{\sigma^2}{k_0}).
$$
So, the set of parameters for the baseline distribution is $(\nu,\lambda,\mu_0,k_0)$. 

\begin{figure}
\centerline{\includegraphics[scale=.5]{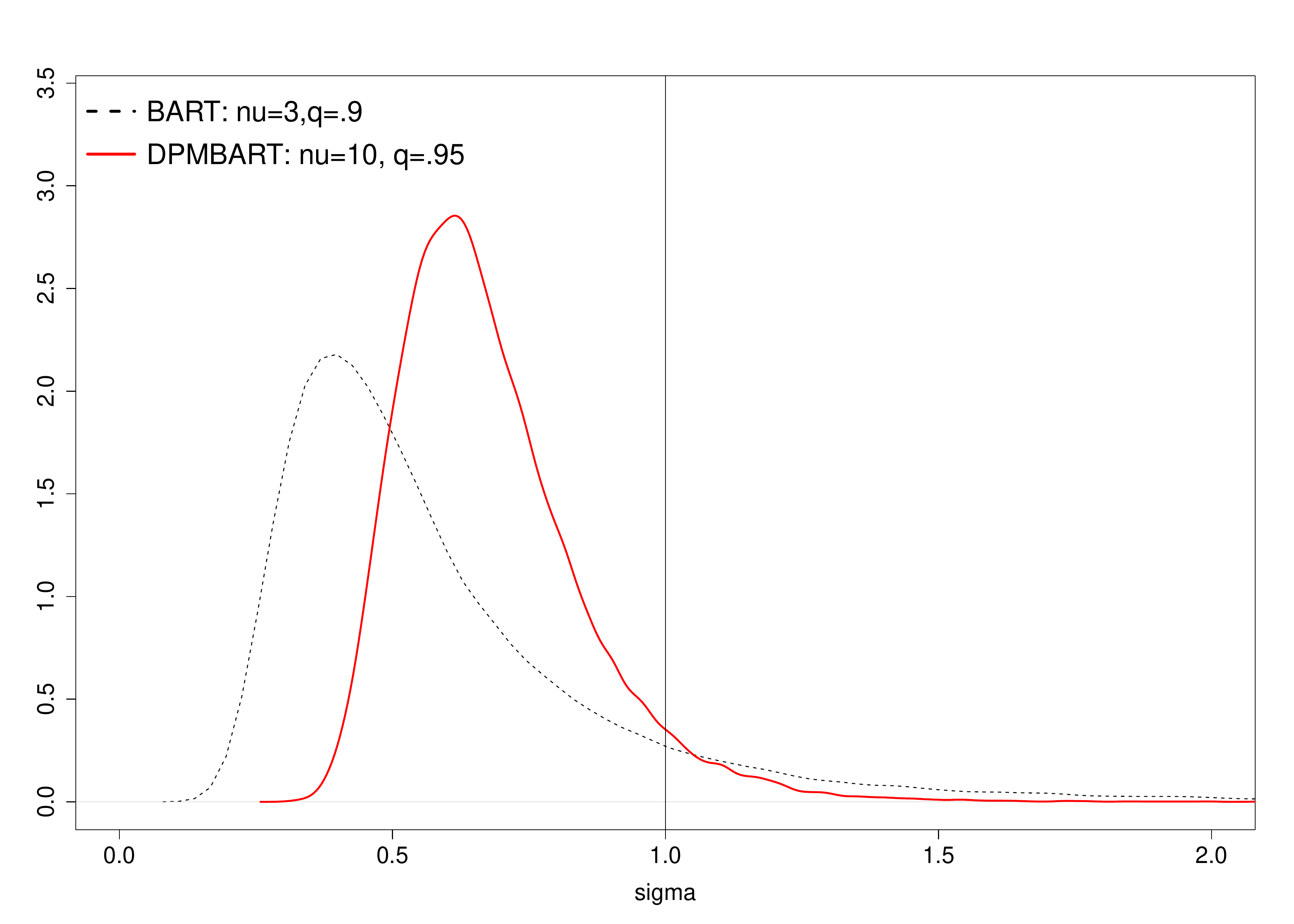}}
\caption{%
Choice of baseline parameter $\lambda$.
For DPMBART we use the same construction as in BART
with $\nu$ increased to 10 and a default quantile of .95.
\label{fig:baseprior-lambda}}
\end{figure}

First, we choose $(\nu,\lambda)$.
We are guided by the BART choice of $(\nu,\lambda)$, but make some adjustments.
First, we can make the prior tighter by increasing $\nu$.
The spread of the error distribution is now covered by having many components so we can make the
distribution of a single component tighter.
In addition, as mentioned above, we do not want to make our baseline distribution too spread out.
Our default value is $\nu=10$ as opposed to the BART choice of $\nu=3$.
We then choose $\lambda$ using the same approach as in BART, but  we increase
the default quantile from .9 to .95.  
With DPMBART we rely on the multiple components to cover the possibility of smaller errors.
Figure~\ref{fig:baseprior-lambda} plots the default DPMBART (solid line) and BART (dashed line) priors for $\sigma$.
The BART prior is forced to use a smaller $\nu$ and quantile to reach the left tail down to cover
the possibility of small $\sigma$.
The DPMBART can be more informative and does not have to reach left.

We now discuss the choice of $(\mu_0,k_0)$.
Typically in application we subract $\bar{y}$ from $y$ which facilitates the relatively
easy choice of $\mu_0=0$.
For $k_0$ we try to follow the same BART philosophy of gauging the prior from the residuals of a linear fit.
For $\sigma$ we gauged the prior using the sample standard deviation of the residuals.
For $k_0$ we again use the residuals but now consider the need to place the $\mu_i$  into the range of 
the residuals.

The marginal distribution of $\mu_i$ given our baseline distribution, is
$$
\mu \sim \frac{\sqrt{\lambda}}{\sqrt{k_0}} \, t_\nu.
$$\sko

We use this result to choose $k_0$ to scale the distribution relative to
the scale of the residuals.
Let $e_i$ be the residuals from the multiple regression.\sko
Let $k_s$ be a scaling for the $\mu$ marginal.\sko
Given $k_s$ we choose $k_0$ by solving:
$$
max |e_i| = k_s \frac{\sqrt{\lambda}}{\sqrt{k_o}}.
$$\sko

The default we use is $k_s=10$.
This may seem like a very large value, but it must be remembered that the conjugate
form of our baseline distribution has the property that bigger $\mu$ go with bigger $\sigma$
so we don't have to reach the $\mu_i$ out to the edge of the error distribution.
It might also make sense to use a quantile of the $e_i$ rather than the max but the results reported
in this paper use the  max.

Note that a common practice in the DPM literature is to put distributions on the parameters.
This is, of course, a common practice  in modern Bayesian analysis and often leads to wonderfully
adaptive and flexible models.
Our attitude here is that the combination of the BART flexibility with the DPM flexibility
is already very adaptable and there is a real premium on keeping things as simple as possible.
Thus, we choose default values for
$(\nu,\lambda,\mu_0,k_0)$ rather then priors for them.
We do, however, put a prior on $\alpha$.

\subsection{Specification of the Prior on $\alpha$}\label{subsec:alphaprior}

The prior on $\alpha$ is exactly the same as in Rossi (see Section 2.5).
The idea of the prior is to relate $\alpha$ to the number of unique components.
The user chooses a minimum and maximum number of components $I_{min}$ and $I_{max}$.
We then solve for $\alpha_{min}$ so that the mode of the consequent distribution for $I$
equals the number of components is $I_{min}$.
Similarly we obtain $\alpha_{max}$ from $I_{max}$.
We then let
$$
p(\alpha) \propto (1 - \frac{\alpha - \alpha_{min}}{\alpha_{max} - \alpha_{min}})^\psi.
$$ 
The default values for $I_{min}$, $I_{max}$, and $\psi$ are 1, $[.1n]$, and .5, where
$[]$ denotes the integer part  and $n$ is the sample size.
The nice thing about this prior is it automatically scales sensibly with $n$.

\subsection{Computational Details}\label{subsec:comp}

Our full parameter space consist of the $\{T_j,M_j\}$ trees, $j=1,2,\ldots,m$,
the $\{\theta_i=(\mu_i,\sigma_i)\}$, $i=1,2,\ldots,n$ and  $\alpha$.

Our Markov Monte Chain Monte Carlo (MCMC) algorithm is the obvious Gibbs sampler:

\begin{eqnarray*}
\{T_j,M_j\} & \C & \{\theta_i=(\mu_i,\sigma_i)\} \\
\{\theta_i=(\mu_i,\sigma_i)\} & \C &  \{T_j,M_j\}\\
\alpha & \C & \{\theta_i=(\mu_i,\sigma_i)\}
\end{eqnarray*}

The first draws relies on the author's C++ code for weighted BART which allows for
$\epsilon_i \sim N(0,w_i^2 \sigma^2)$. The first draw is then easily done using
$Y_i - \mu_i = f(x) + \tilde{\epsilon}_i, \; \tilde{\epsilon}_i \sim N(0,\sigma_i^2)$.
The second draw is just the classic Escobar and West density estimation with $\epsilon_i = Y_i - f(x_i)$.
We just use
draws (a)  and (b)
of the simple algorithm in Section 1.3.3 by Escobar and West in the book by  Dey at al. \cite{DeyEtAl}.
Draw (a) draws all the $\theta_i$, (b) draws all the unique $\theta_i$ given the assignment of shared values.
While there are more sophisticated algorithms in the literature which may work well,
we have had good results using this simple approach.
We wrote relatively simple C++ code to implement it which is crucial because we need a simple
interface for the DPM draws with the C++ code doing the weighted BART draws.

The final draw is done by putting $\alpha$ on a grid and using Baye's theorem with
$p(\alpha \C \{\theta_i\}) = p(\alpha \C I) \propto p(I \C \alpha) \, p(\alpha)$,
where $I$ is the number of unique $\theta_i$.

\section{Examples}\label{sec:examples}
In this section, we present examples to illustrate the inference provided 
by the DPMBART model. 

In Section~\ref{subsec:simex}, we present simulated examples where the errors are drawn
from the t distribution with 20 degrees of freedom,
the t distribution with 3 degrees of freedom,
and the log of a gamma.
The $t_{20}$ distribution  gives us errors which are essentially normal as assumed by BART.
The $t_3$ distribution gives us a heavy-tailed error distributions and some observations
which, in practice, would be deemed outliers.
The log of a gamma gives us a skewed distribution.
These three error distributions cover the kinds of errors we normally think about.
The examples show that when the errors are close to normal, DPMBART provides an error distribution
inference which is very close to BART.
The heavy-tailed and skewed examples show that when the error distribution departs substantially
from normality, there is big difference between DPMBART and BART with DPMBART being much closer
to the truth.
In the non-normal cases the DPMBART error distribution inference comes much closer to uncovering the correct
error distribution but is shrunk somewhat towards the BART inference.
We consider this desirable in that we want to preserve the well established good properties of BART
in the normal case and a little shrinkage may stabilize the estimation in the non-normal case.
In these simulated examples, $x$ is one-dimensional so that we can easily visualize the simulated data
and fit of the function $f$.
In these examples, the estimates of $f$ from BART and DPMBART are very similar.
This is because the signal is fairly strong.  In low signal to noise cases, with substantial outliers,
there can be large differences in the function estimation.

In Section~\ref{subsec:card}, we consider real data with a seven dimensional predictor $x$.
Our data is the first stage regression from the classic Card paper \cite{Card95}  which uses instrumental variables
estimation to infer the effect of schooling on income.
While our analysis is partial in that we only look at the first stage equation,
we easily find that there is substantial nonlinearity and non-normal errors.
Experienced IV investigators must decide if these awareness of these features of the data
should be taken into account in the full IV analysis.

Note that we use the term ``fit'' at $x_i$ from BART or DPMBART  to refer to the value
$$
\hat{f}(x_i) = \frac{1}{D} \sum_{i=}^d \, f^d(x_i)
$$
where $f^d$ is the function obtained from the $d^{th}$ kept MCMC draw of 
$((T_1,M_1),\ldots,(T_m,M_m))$.
We use ``fit'' and $\hat{f}$ interchangeable to refer to the estimated posterior mean.

\subsection{Simulated Examples}\label{subsec:simex}


%

We present results for three simulated scenarios.
In each simulation we have 2,000 observations.
$x_i \sim \text{Uniform}(-1,1)$.
The function $f$ is $f(x) = 10 \, x^3$.
In our first simulation, $\epsilon_i \sim t_{20}$, the t-distribution with 20 degrees of freedom.
In our second simulation, $\epsilon_i \sim t_3$, the t-distribution with 3 degrees of freedom.
In our third simulation,  we
(i) generate $\epsilon_i \sim \text{Gamma}$, with the shape parameter set to .3 (see \texttt{rgamma} in \texttt{R});
(ii) flip the errors: $\epsilon_i \rightarrow -\epsilon_i$;
(iii) demean the errors: $\epsilon_i \rightarrow \epsilon_i - \bar{\epsilon}$.
In each case, the draws are iid.

\begin{figure}
\centerline{\includegraphics[scale=.6]{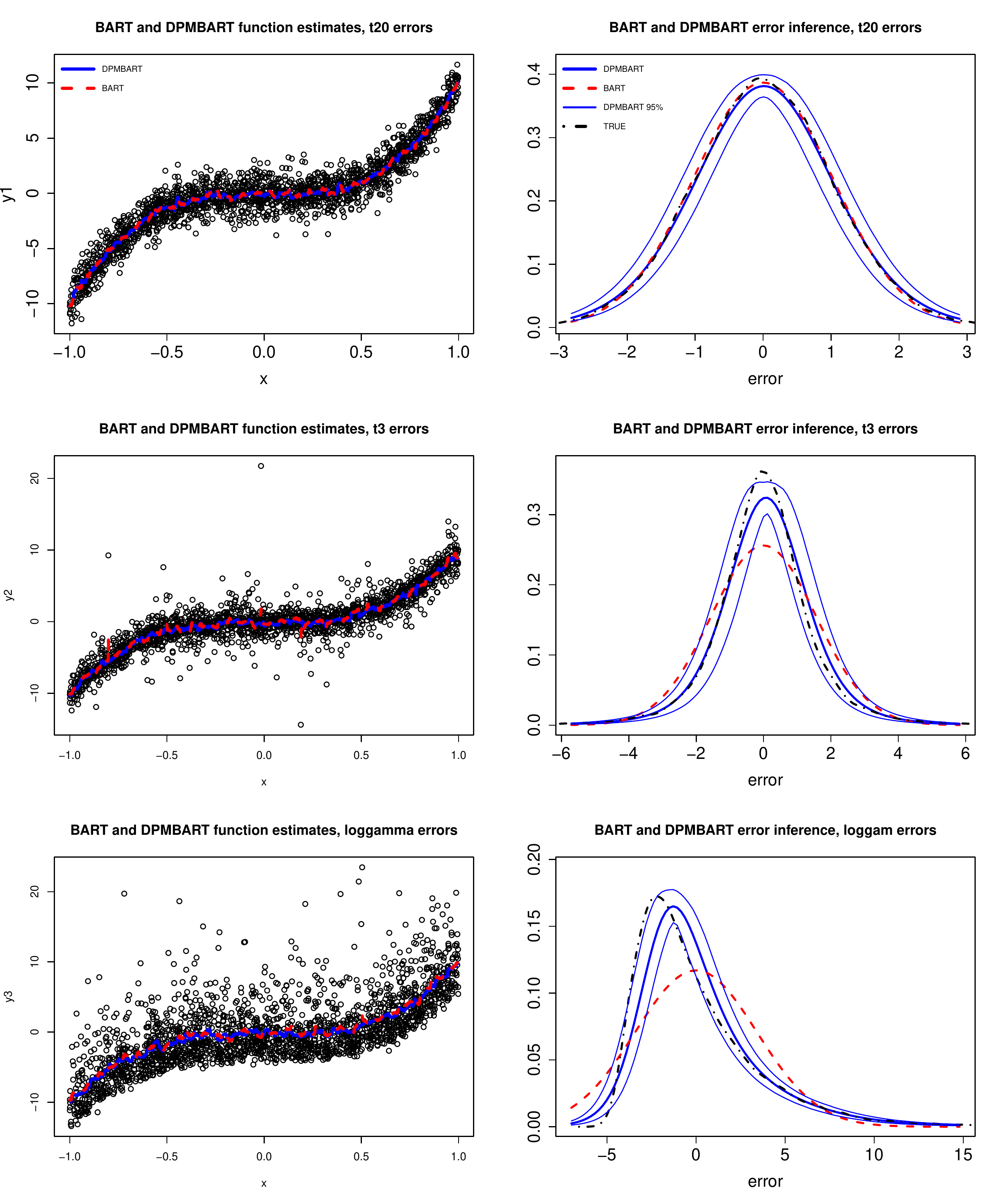}}
\caption{%
Function and error distribution estimation for three simulated examples.
The three rows correspond to the three error distributions.
The first column shows the simulated data and function estimation.
The second column shows the error distribution inference.
\label{fig:plot-3sims}}
\end{figure}

Figure~\ref{fig:plot-3sims} displays the results for a single simulation from each of our three errors distributions.
The three rows of the figure correspond to the three error distributions.
The first column displays the simulated data and the estimates of $f$ based on
the posterior means of $f(x)$ for each $x$ in the training data.
For example, in the (2,1) plot, we can clearly see the heavy tail and apparent outliers.
In the (3,1) plot, the strong skewness of the errors is evident.
The function estimates from DPMBART are plotted with a solid line (blue) and the estimates from BART
are plotted with a dashed line (red).
With the strong signal, there is little difference in the function estimates.

The second column displays our inference for the error distribution.
The solid thick line (blue) displays the average of the density estimates from DPMBART MCMC draws
(which is the predictive distribution of an error).
The thin solid lines (blue) show 95\% pointwise intervals for the true density based on the MCMC draws.
The dashed line (red) shows the BART average normal error distribution, averaged over MCMC draws of $\sigma$.
The dot-dash line (black) shows the true error distribution.
In the (1,2) plot we see that when the errors are essentially normal, the difference between the BART
and DPMBART error inference is very small with little uncertainty.
In both the (2,2) (heavy tails) and the (3,2) (skewness) cases we see that DPMBART estimation is much
closer to the truth than the BART estimation.  The true density is {\it almost} covered everywhere by
the 95\% intervals.
The DPMBART estimation seems to be pulled slightly towards the BART estimate, which is what we want.

Figure~\ref{fig:plot-3sims} succinctly captures the spirit of the paper.
Of course, we cannot be sure that DPMBART will perform as well in all situations, in particular,
in higher dimensional situations, but so far it looks promising.

\begin{figure}
\centerline{\includegraphics[scale=.6]{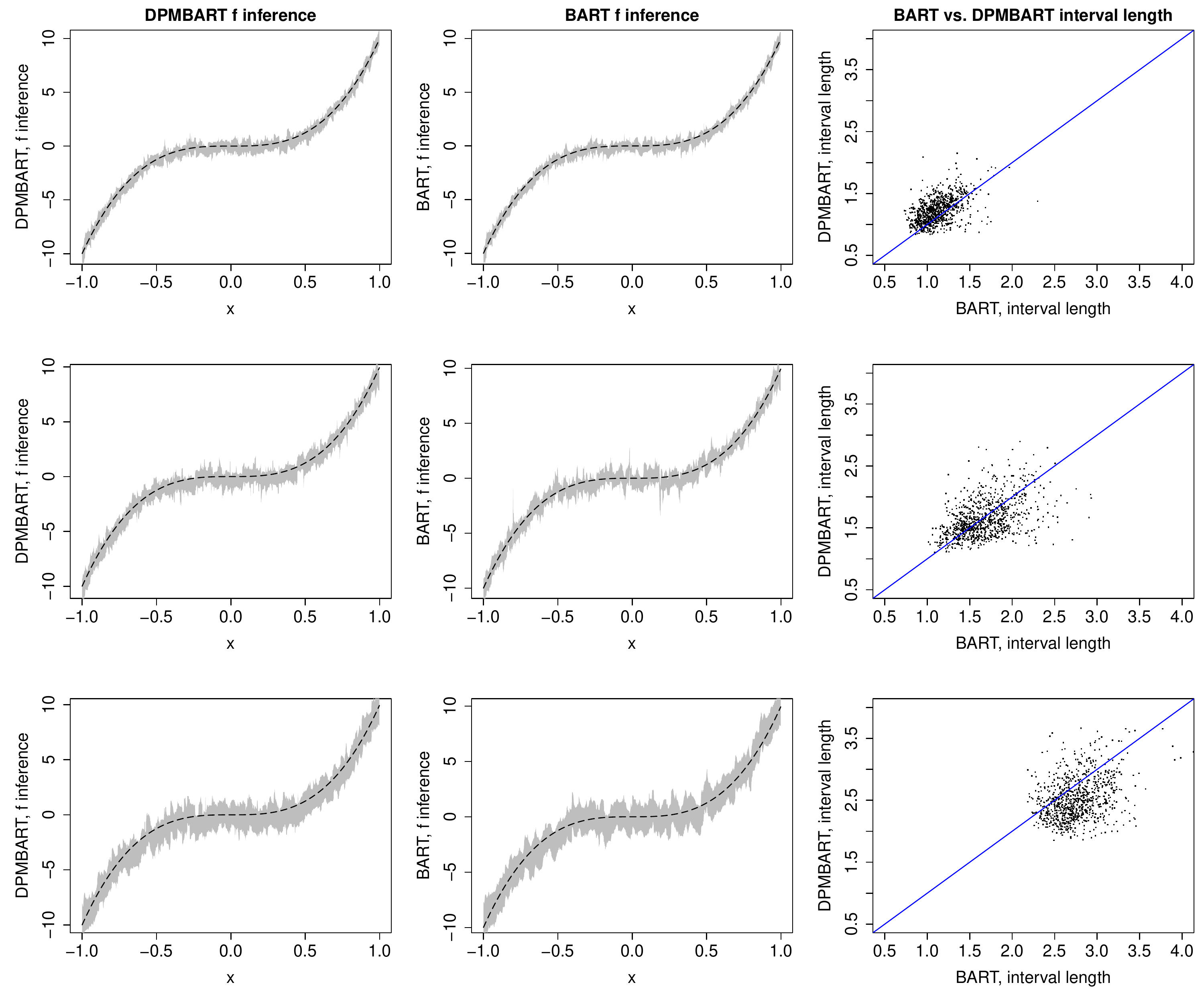}}
\caption{%
As in Figure~\ref{fig:plot-3sims}, the three rows correspond to our three error distributions.
95\% intervals for the  $f(x_i)$ from DPMBART (first column) and BART (second column).
The third column plot the width of the intervals from BART versus with widths from DPMBART.
\label{fig:sim-f-inference}}
\end{figure}

Figure~\ref{fig:sim-f-inference} illustrates BART and DPMBART 
uncertainty quantification for inference of the function $f$.
As in Figure~\ref{fig:plot-3sims}, the three rows correspond to our three error distributions.
In the first column we plot point-wise 95\% posterior intervals for $f(x_i)$ vs $x_i$ from DPMBART.
The second column displays the 95\% intervals from BART.
In each plot the true $f$ is plotted with a dashed line.
In the third column we plot the width of the interval for each $f(x_i)$ from BART versus
the corresponding quantity from DPMBART.
We see that according to both BART and DPMBART there is considerably more uncertainty with
the log gamma errors which is plausible given the plots of the data in the first column of
Figure~\ref{fig:plot-3sims}.
For the $t_{20}$ errors, the DPMBART intervals tend to be large which makes sense
since when the errors are approximately normal, BART should know more.
For the log gamma errors, the BART intervals are large, which makes sense because 
DPMBART can figure out that BART is wrong to assume normal-like errors.

\subsection{Card Data}\label{subsec:card}

In a famous paper,
Card uses instrumental variables to estimate the returns to education.
A standard specification of the first stage regression relates the treatment variable 
years-of-schooling (\texttt{ed76}) 
to measures of how close a subject lives to a two and a four year college (\texttt{nearc2}, \texttt{nearc4}),
experience and experience-squared, a race indicator (\texttt{black}),
an indicator for whether the subject lives in a standard metropolitan area (\texttt{smsa76r}),
and an indicator for whether the subject lives in the south (\texttt{reg76r}).
The measures of proximity to college are the instruments in that they plausibly induce exogenous variation
in the cost of education and hence the amount of education.
The second stage equation relates wages to the years of schooling.

Rather than looking at the full system given by both equations, we explore
a partial analysis by just considering estimation of the first stage equation
which is generally considered to follow the framework of our DPMBART model
$Y = f(x) + \epsilon$.
Most analyses assume $f$ is linear and most Bayesian analyses assume normal errors with
the notable exception of \cite{CHMR08} which assumes linearity but uses a bivariate DPM to model the two
errors in the first and second stage equations.

Below is the standard output (in \texttt{R}) from a linear multiple regression of 
Y=treatment=years-of-schooling on the explanatory variables and instruments.

\begin{samepage}
\begin{nobreak}
{\small
\begin{verbatim}
Coefficients:
             Estimate Std. Error t value Pr(>|t|)
(Intercept) 16.566718   0.121558 136.287  < 2e-16 ***
nearc2       0.107658   0.072895   1.477 0.139809
nearc4       0.331239   0.082587   4.011 6.20e-05 ***
exp76       -0.397082   0.009224 -43.047  < 2e-16 ***
exp762       0.069560   0.164981   0.422 0.673329
black       -1.012980   0.089747 -11.287  < 2e-16 ***
smsa76r      0.388661   0.085494   4.546 5.68e-06 ***
reg76r      -0.278657   0.079682  -3.497 0.000477 ***
---
Signif. codes:  0 ‘***’ 0.001 ‘**’ 0.01 ‘*’ 0.05 ‘.’ 0.1 ‘ ’ 1

Residual standard error: 1.942 on 3002 degrees of freedom
Multiple R-squared:  0.4748,    Adjusted R-squared:  0.4736
F-statistic: 387.8 on 7 and 3002 DF,  p-value: < 2.2e-16
\end{verbatim}
}
\end{nobreak}
\end{samepage}

\begin{figure}
\centerline{\includegraphics[scale=.6]{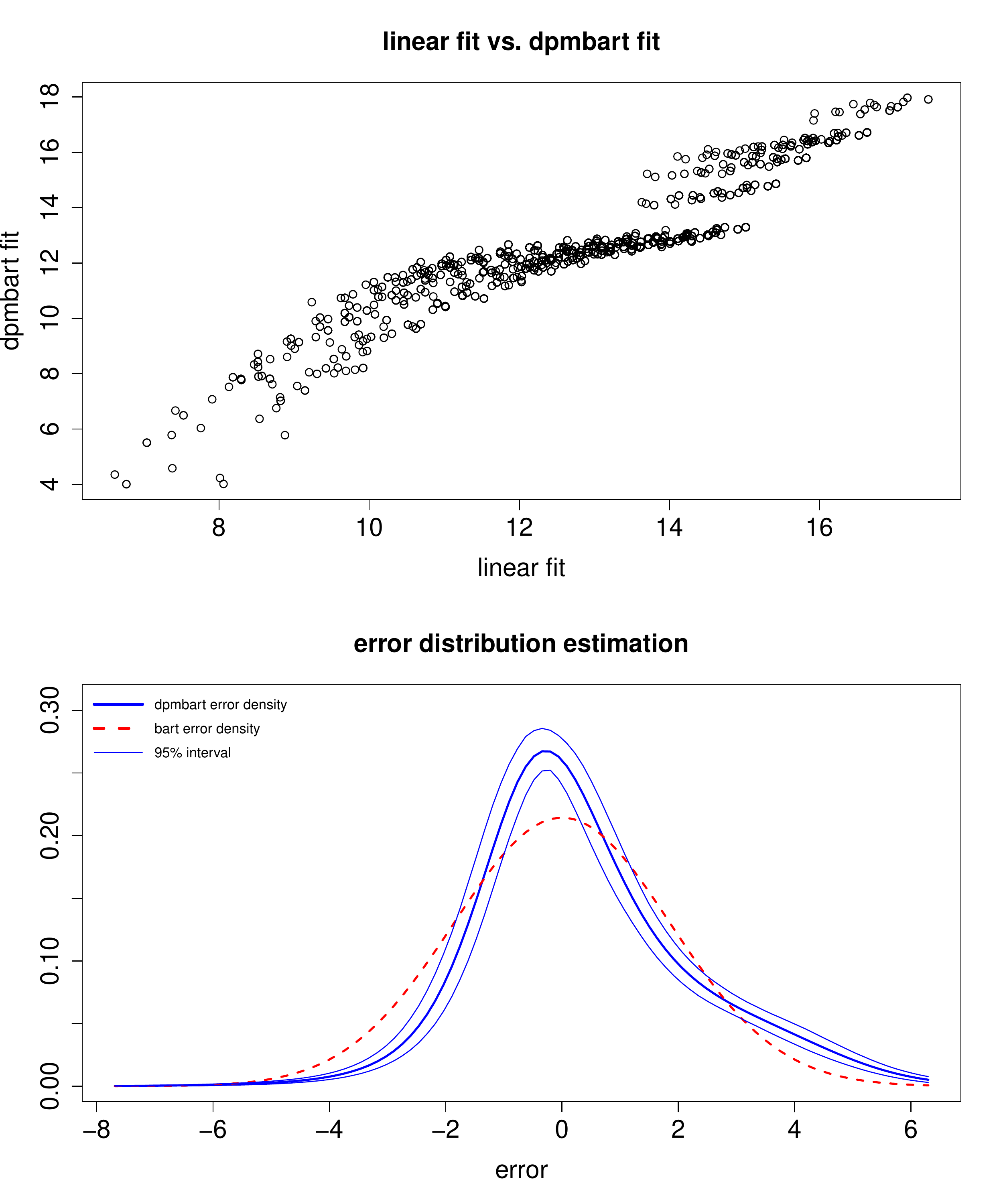}}
\caption{%
Top panel: fitted values from a linear regression versus the fitted values from DPMBART.
Bottom panel: DPMBART and BART error distribution estimation.
\label{fig:card-data-dpmbart}}
\end{figure}

The top panel of Figure~\ref{fig:card-data-dpmbart} plots the fitted values from the  linear regression versus
the fitted values from DPMBART.
Clearly,  DPMBART is uncovering non-linearities that the linear model cannot.
However the R-squared ($\text{cor}(y,\hat{f}(x))$) from the DPMBART is .5 which is not appreciably different
than the linear regression R-squared of .48 reported above.
The bottom panel of Figure~\ref{fig:card-data-dpmbart} displays the error distribution inference
using the same format as Figure~\ref{fig:plot-3sims} except that there is no known true error density.
Our DPMBART model suggests strong evidence for heavy tailed and skewed errors.

\begin{figure}
\centerline{\includegraphics[scale=.6]{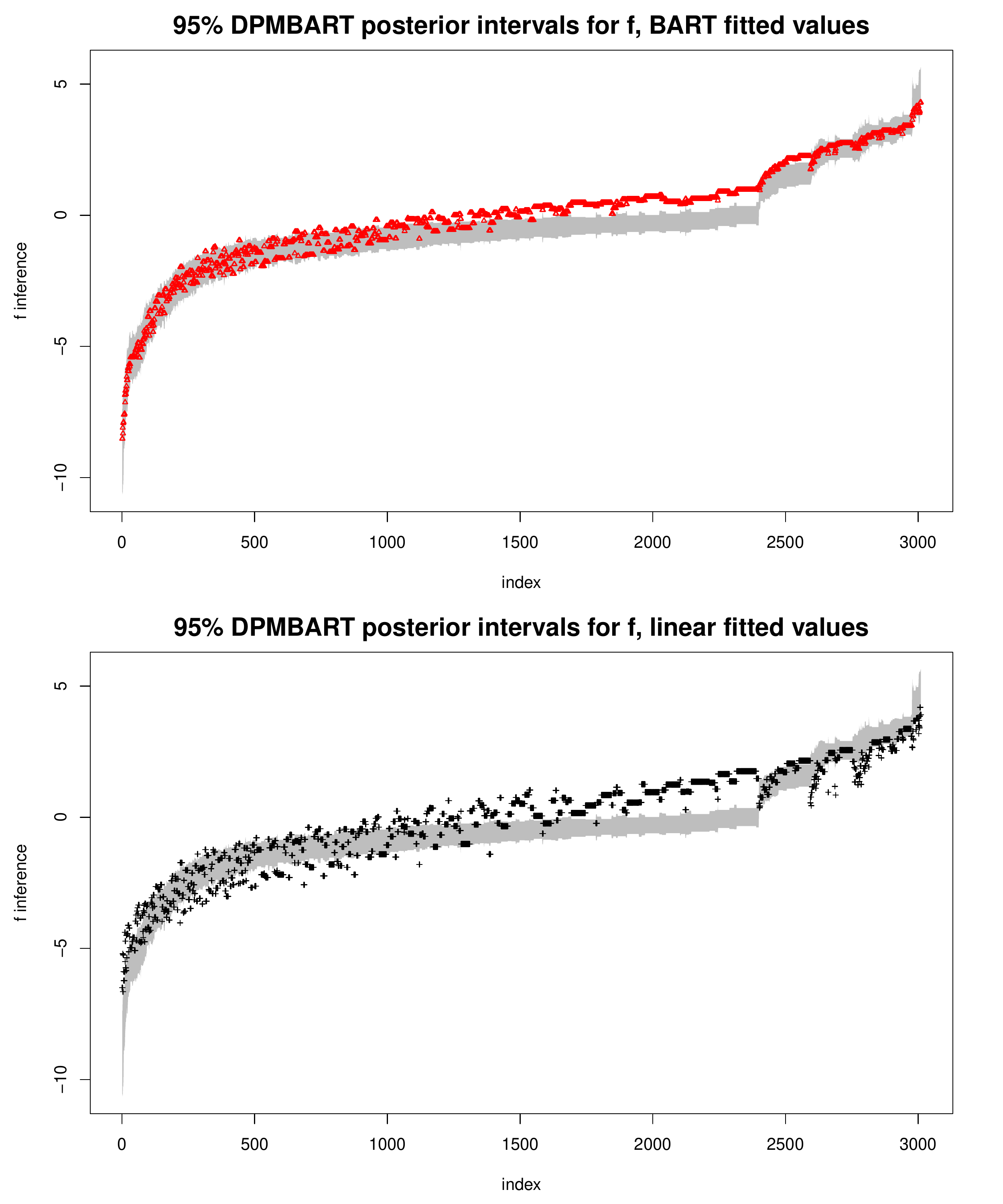}}
\caption{%
DPMBART function inference and fits from BART (top panel)
and linear regression (bottom panel).
\label{fig:card-f-inference}}
\end{figure}

Figure~\ref{fig:card-data-dpmbart} suggests that DPMBART uncovers features of the function $f$
not visible to the linear model.
To assess this further and quantify our estimation uncertainty, we plot 95\% posterior intervals
for $f(x_i)$ from DPMBART for each $x_i$.  We sort the intervals by the value of $\hat{f}(x_i)$.
In the top panel of Figure~\ref{fig:card-f-inference} we display the DPMBART 95\% intervals and
use points (triangles) to plot the fits from BART.
In the bottom plot we again plot the DPMBART intervals, but use symbols (+) to plot the fitted values
from linear regression.
There appears to be a set of observations where (sorted index values 1500-2500) the fitted values
from BART and the linear model are well outside the DPMBART posterior intervals.
This suggests that there is ``statistically significant'' evidence that the methods have different views
about $f$.  Whether these differences are of practical importance for the final causal inference is now
as open question.

\begin{figure}
\centerline{\includegraphics[scale=.6]{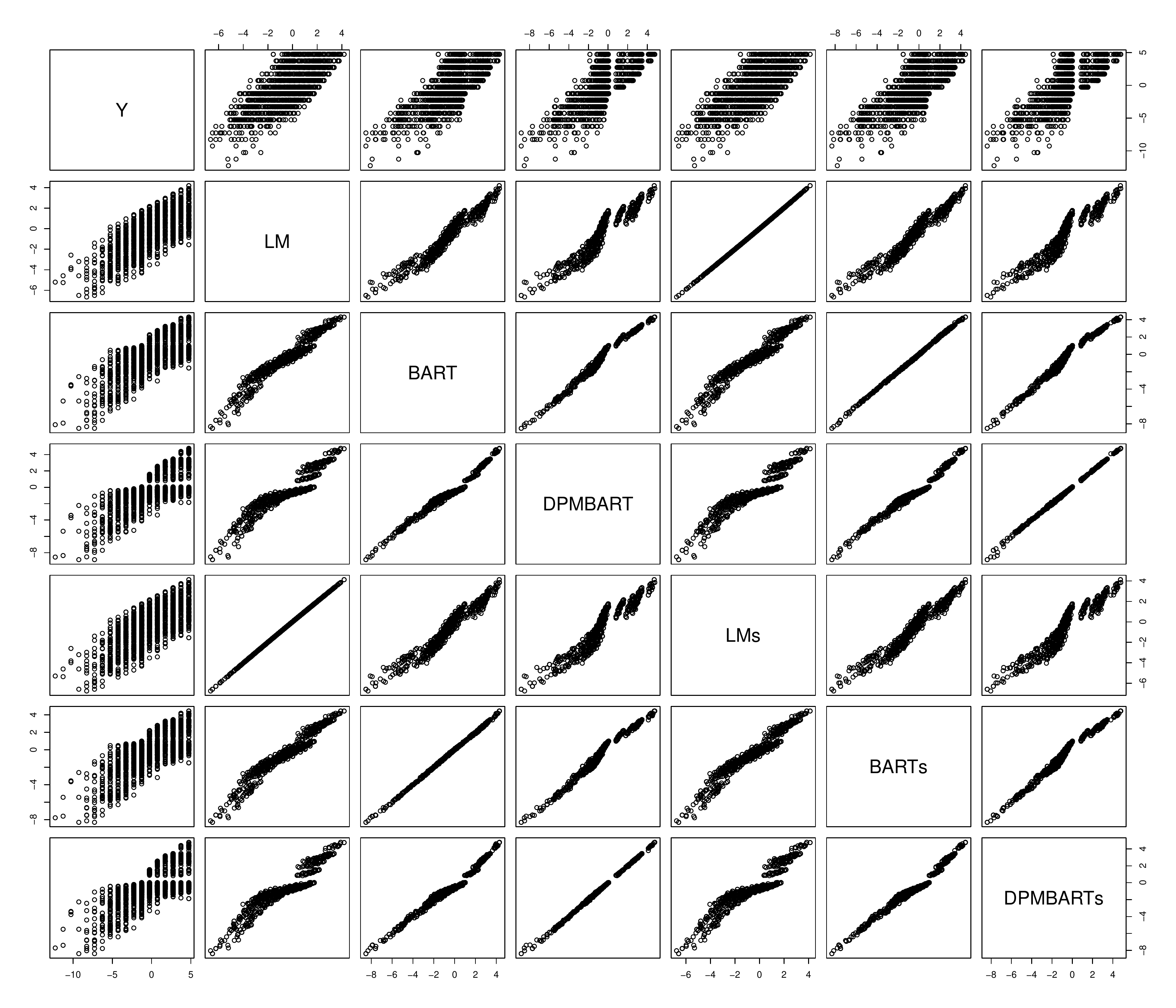}}
\caption{%
Dependent variable Y=treatment=years-of-schooling and various in-sample fits.
LM, BART, DPMBART are the in sample fits from a linear model, BART, and DPMBART.
The same names with the with a small ``s'' appended are fits from the same methodology
but without experience-squared included.
\label{fig:card-data-dpmbart-pairs}}
\end{figure}

In Figure~\ref{fig:card-data-dpmbart-pairs}, $Y$ and the fitted values from the linear model,
DPMBART, and BART, as well was the fitted values from the same three methods excluding
the regressor experience-squared. The fits excluding the square have ``s'' appended to the name.
So, for example, BARTs, is the fits from running BART without experience-squared.
From the top row we see that the six fits are not dramatically different in their fit to $Y$.
All models give very similar fits when experience-squared is excluded.
Note that a basic appeal of BART is that there is no need to explicitly include a transformation of an
explanatory variable. 
As in Figure~\ref{fig:card-data-dpmbart} the DPMBART fit is noticeable different from the linear fit.
Unlike our simulated examples, the BART fits and DPMBART fits are different enough that the convey
quite different messages, in this case about the adequacy of the linear fit. 

\begin{figure}
\centerline{\includegraphics[scale=.6]{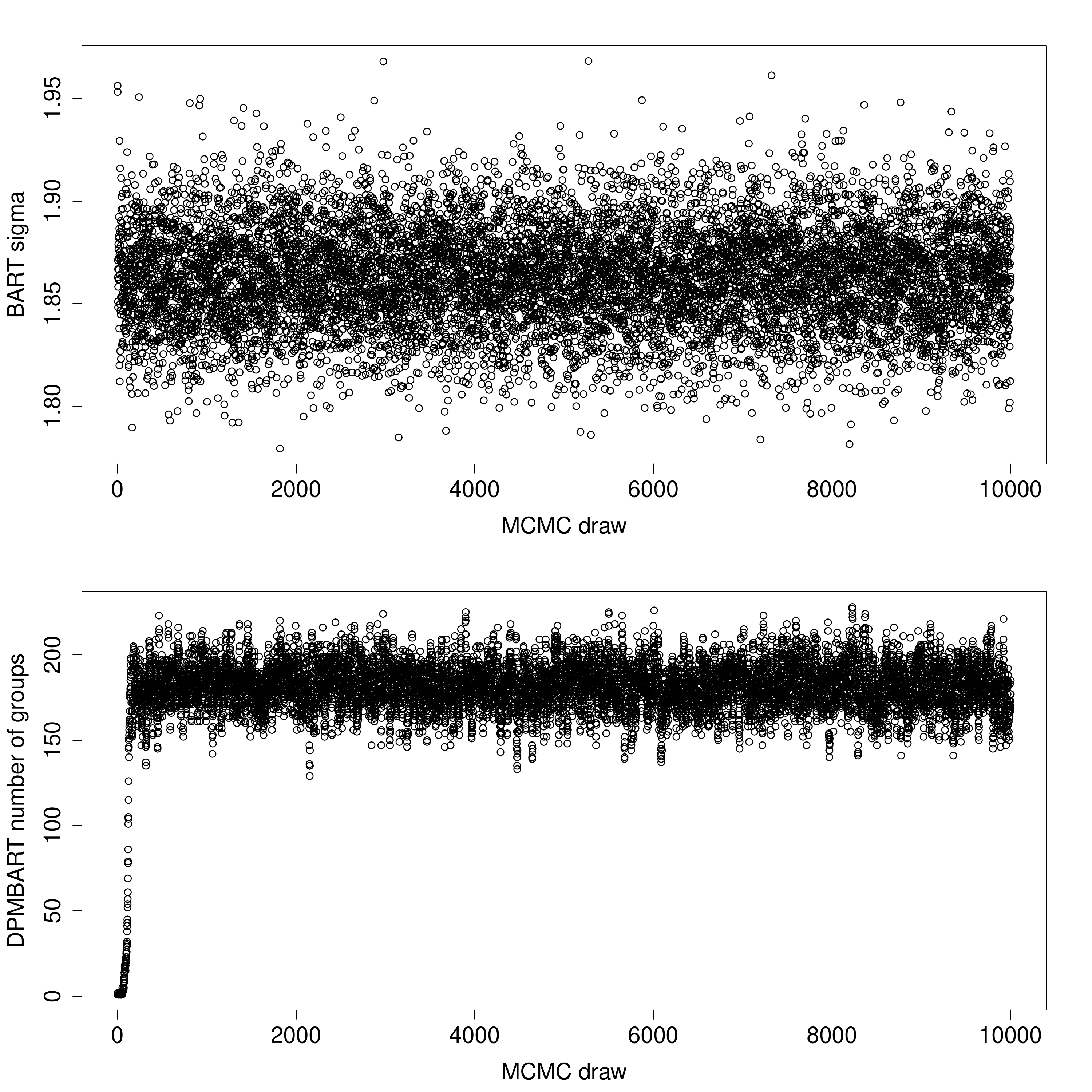}}
\caption{%
MCMC burn in.
Top panel: $\sigma$ draws from BART.
Bottom panel: number of distinct $\theta=(\mu,\sigma)$ from DPMBART.
\label{fig:card-burn-in}}
\end{figure}

Figure~\ref{fig:card-burn-in} gives us a look at the BART and DPMBART MCMC's.
In both cases we ran the MCMC for 10,000 iterations and the first 5,000 were deemed
``burn-in'' and all reported results are from the second set of 5,000 iterations.
The top panel of Figure\ref{fig:card-burn-in} displays all 10,000 draws of $\sigma$.
The draws seem to burn-in immediately.  BART is not always this good!!
The bottom panel displays the number of unique $\theta_i$.
The sampler was started with just one unique value which we recorded and then we recorded
the number after each of the 10,000 MCMC iterations so there are 10,001 values.
So the values start at one, and then very quickly move up to a steady state varying about 180.

\newpage
\section{Conclusion}\label{sec:conclusion}
DPMBART is a substantial advance over BART in that the highly restrictive
and unrealistic assumption of normal errors is relaxed.
Our Bayesian ensemble modeling is now fully nonparametric.
A model with both flexible fitting of the response function $f$
and the error distribution has the potential to 
uncover the essential patterns of the data
without making strong assumptions.
In addition, our default prior seems to give reasonable results without
requiring a great deal of tuning on the part of the user.
By keeping things simple and using intuition from both BART and
the Dirichlet process mixture model we feel we have found an approach
with the potential to deliver a tool of that can be used 
reasonably easily in many applications.

Figure~\ref{fig:plot-3sims} illustrates how DPMBART gives inference similar
to BART when the errors are close to normal but gives a much better inference when
they are not.
Of course, our approach is based on informal choices of key prior parameters
and only further experience will show if the DPMBART prior is sufficiently robust to
work well in a wide variety of applications.

In addition, the nature of the function $f$ is the same in BART and DPMBART
so many of the approaches for understanding the inference for $f$ that have
been used in the past apply.
However, using BART also entails looking at the $\sigma$ draws and new approaches
are needed to understand the more complex inference of DPMBART.

\newpage
\bibliographystyle{plain}
\bibliography{dpmbart}

\end{document}